\documentclass[a4paper, 10pt, conference]{ieeeconf}      

\IEEEoverridecommandlockouts                              
\usepackage{graphics} 
\usepackage{algpseudocode}
\usepackage[linesnumbered,ruled,vlined]{algorithm2e}

\title{\LARGE \bf
How you see me 
}

\author{Rohit Gandikota$^{1}$ and Deepak Mishra$^{2}$
\thanks{$^{1}$Rohit Gandikota is a graduate student from Avionics department,
        Indian Institute of Space science and Technology, Kerala, India
        {\tt\small gandikota.SC15B088 at ug.iist.ac.in}}%
\thanks{$^{2}$Deepak Mishra is with Faculty of Avionics, Indian Institute of Space science and Technology, Kerala, India
        {\tt\small deepak.mishra at iist.ac.in}}%
}
\makeatletter

\newcommand{\Rmnum}[1]{\expandafter\@slowromancap\romannumeral #1@}
\makeatother

\begin{document}

\maketitle
\thispagestyle{empty}
\pagestyle{empty}

\begin{abstract}

Convolution Neural networks(CNN) are one of the most powerful tools in the present era of science. There has been a lot of research done to improve their performance and robustness while their internal working was left unexplored to much extent. They are often defined as black boxes that can map non-linear data effectively. This paper tries to show how we have taught the CNN's to look at an image. Visual results are shown to explain what a CNN is looking at in an image. 
\par The proposed algorithm exploits the basic math behind a CNN to backtrack the important pixels. This is a generic approach which can be applied to any network till VGG. This doesn't require any additional training or architectural changes. In literature, few attempts have been made to explain how learning happens in CNN internally, by exploiting the convolution filter maps. This is a simple algorithm as it does not involve any cost functions, filter exploitation, gradient calculations or probability scores. Further we demonstrate that the proposed scheme can be used in some important Computer Vision tasks.

\end{abstract}

\section{INTRODUCTION}

Convolution Neural networks have been revolutionizing the area of Computer vision with their outstanding performance in vision tasks. They 
are non linear functions which are designed to model a human eye. Acknowledging their importance, research has been done to improve their performance. Over time, the accuracy of the system increased and so did the complexity behind it's working. Much complex architectures are being introduced to improve their performance. For example, starting from AlexNet\cite{c1} with 8 layers, then came ZFNet\cite{c2} and VGGNet\cite{c3}. Present Resnet\cite{c4} has hundreds of layers with 50 times lesser number of parameters compared to AlexNet. These models offer less evidence on how they work internally and achieve such extraordinary results. 
\par One approach for understanding CNN is by exploiting the feature activation maps of the filters in the network. Another way is by proposing the regions that the CNN is looking at, in the image. The common motivation behind these methods are to propose the regions in the image that corresponds to the CNN's output of recognizing objects in the image. Understanding the working of CNN's are important because

\begin{itemize}

\item We can guide the training the network more accurately when we can visualize how it is learning from each epoch of training.
\item A visual evidence can be given to explain any alternate identification(i.e A different classification from the ground truth). Some times a network can identify or move it's attention more towards another less important object in an image. This can be visualized also.
\item Development in the area of neurology can happen by understanding from the CNN's working which is nothing but a human eye model. 
\item Many other applications can be improved and simplified like computing image saliency and improvising detection methods. Self driving cars can use this as they can understand what human sees on the road before he takes certain decisions.

\end{itemize}
 \begin{figure}[thpb]
      \centering
       \includegraphics{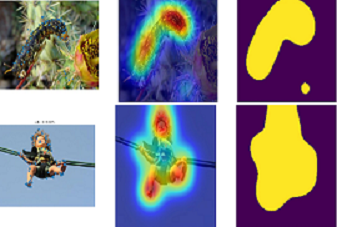}
      \caption{Results from MSRA-B dataset. First column shows the important pixels, second column shows the attention regions and the final column depicts the saliency region in the images.}      
        \label{figurelabel}
   \end{figure}
The proposed algorithm basically exploits the basic working of the CNN to backtrack and find the important pixels in the image, the CNN is looking at to make a decision. We try to unroll the forward pass in the CNN after passing an image, given a node we try to find the nodes that activate the given node the most. This has been done from the final output layer till the input layer to get the information in pixel level. So given an image we pass it through a CNN to get the recognition output and then we simply backtrack the nodes in a tree fashion from the output till the input layer.\par
Rest of the paper is organized as follows: section \Rmnum{2} discusses the existing works that are relevant for the work, section \Rmnum{3} presents the proposed method in detail, section \Rmnum{4} explains the potency of our method on saliency detection empirically, section \Rmnum{5} talks about the future work and improvements we are planning on and finally section \Rmnum{6} concludes the work presented.  
\section{PREVIOUS WORK}
In this sections previous works in this area of research are discussed. This work is mainly motivated by a very dynamic algorithm called Viterbi Algorithm\cite{c11}. Viterbi algorithm is for finding optimal sequence of hidden states. Given an observation sequence and a Hidden Markov Model(HMM), the algorithm returns the state path through the HMM that has a maximum likelihood for the observations. Viterbi algorithm is similar to a forward pass, except has one component: back-pointers. The reason is that while the forward algorithm needs to produce an observation likelihood, the Viterbi algorithm must produce a probability and the most likely state sequence. We compute this sequence by keeping a check on the path of hidden states that led to each state, and then at the end backtracking the best path to the beginning (the Viterbi backtrack). \par Few attempts were made to understand CNN previously, most of them are  gradient based. They find out the image regions which can improve the probability score for a output class. The work presented in \cite{c5} measures the sensitivity of the classification score for a class by tweaking the pixel values. They compute partial derivative of the score in the pixel space and visualize them. Another approach was visualizing using deconvolution, as shown in \cite{c6}. The deconvolution approach visualizes the activation maps(filter maps) learned by the network.
\par Some other works like \cite{c7},\cite{c8},\cite{c10} were done by taking relevance score for each feature with respect to a class. The idea was to see how the output was effected if a feature is dropped. The importance of the feature was based on the change in the score.

\section{PROPOSED METHOD}
This sections discusses the proposed method and it's algorithm in detail. Our work is based on Convolutional neural network, and hence understanding it is vital. Every CNN is made of common blocks in the form of convolution, pooling and fully connected. In this section we explain how we back tracked through these layers to determine discriminative image regions in the form of important pixels. 
\par There is no need for any extra training for this, we used a pre-trained VGG19 network on Imagenet dataset.Our algorithm exploits the CNN working during testing time. Backtracking fully connected, convolution and pooling layers are described in detail below

\begin{figure}[thpb]
      \centering
       \includegraphics{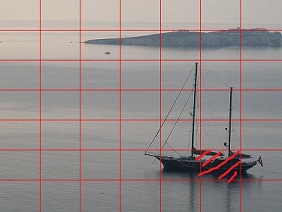}
      \caption{After backtracking through all the fully connected layers, we end up with nodes in 7x7x512 dimensional space. Projecting the top responsible node in that dimension onto the input image of size 224x224 results in the above visualization. The image has been split into equal 7x7 boxes and box with same co-ordinates as the node is selected to visualize the important region.}
        \label{figurelabel}
   \end{figure}
\subsection{Backtracking Fully Connected Layers}Typically for recognition the final layer will be a fully connected with a soft-max activation. We maintained a two memory vectors, \(M1\) for storing the important node locations in the succeeding layer and \(M2\) to start storing the locations of important nodes in the present layer. For every node \(N\) in the memory vector \(M1\), we track the nodes in the present layer that are most responsible to activate it. So for example in the last layer since we are interested in the predicted label, our memory vector \(M1\) has only one node corresponding to the label . We start storing all the nodes in the fc2 layer of VGG19 that are most responsible(i.e. top \('n'\) number of nodes that contribute) for the activation of the node in memory vector \(M2\). Fig. 3 shows the process of backtracking through fc layers.  \par For the of understanding \textit{Algorithm 1}, let us assume that the memory vector of succeeding layer has m number of nodes. And since the fc layers are \( 1\) dimensional, we stored the locations by just a single number. But for the first fc layer whose previous layer would be spatial layer, we need to store the location in a tuple of 3 dimensions \newline \([filter, x, y]\). As shown in Fig.3. This tuple can be visualized over an image by plotting the top responsible node on the image as a box. Further we backtrack through convolution and max pooling layers till the input image to plot important pixels on the image.
\begin{algorithm}
\caption{Backtracking through fully connected layers}
\SetAlgoLined
\DontPrintSemicolon
M1 : Memory vector that has node locations from higher layer\;
M2 : Empty memory vector to store locations of present layer\;
m: Number of nodes in M1 vector\;
\For{i=1:m}{
W,b=weights and bias from this layer to node M1[i]\;
A = Activations in the present layer\;
array = W*A + b\; 
M2.append(arg(array \(>\) 0))
}
M1 \(<\)- M2
\end{algorithm}

\subsection{Backtracking through Convolution Layers}
As discussed above upon reaching the first fc layer while backtracking, the next layer would be convolution or pooling layers. This subsection would talk about how we back tracked through these layers. Note that from here, all the layers have a 2D(in case of pooling) or 3D(in case of convolution) receptive fields. Also we have stored our nodes in the first fc layer in 3d tuple. \par
As shown in Fig. 4, for each important node in the present layer \(l\), we extract the receptive field from the previous layer \(l-1\). Now we calculate the dot product between weights and check for the feature in \(l-1\) layer that has maximum activation. Later to get the x and y co-ordinates, we took the node with maximum activation in the feature map that we have extracted earlier. Note that once the dot product is calculated, the result would be the same shape as receptive field in the previous layer. We calculated the sum over x and y axes to find out the maximum activating feature. Similarly after we extracted the feature, we just took the maximum activating node in that feature. This location is stored in a memory vector that is sent back to above layers. Algorithm 2 explains the above mentioned process

\begin{algorithm}
\caption{Backtracking Convolution layers}
\SetAlgoLined
\DontPrintSemicolon
M1 : Memory vector that has node locations from higher layer\;
M2 : Empty memory vector to store locations of present layer\;
\For{i=1:m}{
W,b=weights and bias from this layer to node M1[i]\;
A = Activations in the present layer\;
array = W*A + b\; 
C = sum(array,axes=x,y)\;
channel = arg(C \(>\) 0)\;
x,y = unravel\_index(argmax(array[channel]))\;
M2.append([channel,x,y])\;
}
\end{algorithm}

\subsection{Backtracking through Max pooling Layers}
As discussed earlier, Max pooling has a 2D receptive field. As explained in \textit{Algorithm 3} and visualized in Fig.5, for every important node in the present layer, we extract the receptive field in the previous layer and find the node with maximum activation. 
\begin{algorithm}
\caption{Backtracking Maxpooling layers}
\SetAlgoLined
\DontPrintSemicolon
\(rf(n)\) = function that extracts receptive field of the node \(n\)
M1 : Memory vector that has node locations from higher layer\;
M2 : Empty memory vector to store locations of present layer\;
\For{i=1:m}{
A = Activations in the present layer\;
array = \(rf(M1[i])\)\; 
C = argmax(array)\;
channel = M1[i][0]\;
x,y = unravel\_index(C)\;
M2.append([channel,x,y])\;
}
\end{algorithm}
\par Note that we have basically unwrapped the working of all the layers to backtrack the important pixels in the image. So after passing the image through the CNN, we extract the activation maps and go backwards unwrapping all the layer functions and finally reach the input layers. That after visualizing gives the important pixels in the image, as shown in Fig.9.

   \begin{figure}[thpb]
      \centering
      \includegraphics{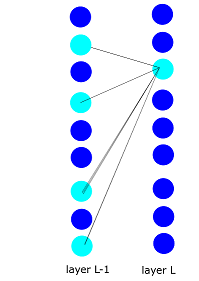}
      \caption{Backtracking through fully connected layers. The turquoise colored nodes in layer \( L-1\) positively activate the node in layer \(L\). These are the nodes that are stored in the memory \(M2\) }
      \label{figurelabel}
   \end{figure}
   
     \begin{figure}[thpb]
      \centering
      \includegraphics{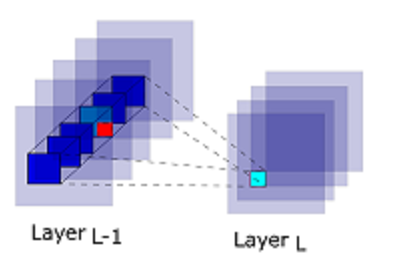}
      \caption{Backtracking through Convolution layers. The turquoise colored channel in layer \( L-1\) is the maximum contributing channel for the activation of turquoise colored node in layer \(L\). Further the red colored node is selected as it is the maximum contributing node from the selected channel. This channel and the co-ordinates of the red colored node are stored in the memory \(M2\)}
      \label{figurelabel}
   \end{figure}
     \begin{figure}[thpb]
      \centering
      \includegraphics{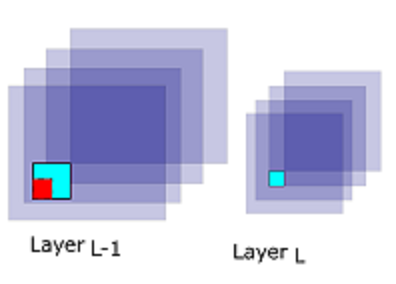}
      \caption{Backtracking through Max pooling layers. The turquoise colored 2x2 nodal region in layer \( L-1\) is the receptive field for max pooling operation of the node highlighted in layer \( L\). Further the red colored node is the maximum contributing node from the field. This channel and the co-ordinates of the red colored node are stored in the memory \(M2\)}
      \label{figurelabel}
   \end{figure}

\section{APPLICATIONS}
There are many applications associated with this method. For example, it can be used as a visual tool for attention regions of CNN in the image. This method can be used a object detection by drawing bounding boxes around the extrema important pixels. Also to detect saliency regions in the image. This can also be used for better training and understanding of CNN. Note that the simplicity in the method makes the computational complexity very less for all the above mentioned applications. We have done some detailed analysis on saliency detection.   
\subsection{Saliency detection}
After obtaining the important pixels in the image as shown in Fig.6, we have drawn Gaussian around each pixel and thresholded to obtain the saliency maps as shown in Fig.7. We have ran our algorithm on MSRA-B dataset and the results are shown in Table 1. Attention region can also be visualized without thresholding the Gaussians. Best values were chosen for standard deviation of Gaussian and the threshold value after considering several random values. Some interesting observations were seen, like
\begin{itemize}
\item Even though the class of the object present in the image is not known to CNN, it looks at the object at meaningful regions.
\item This proposed method works good with blurred images also as shown in Fig.10(a).
\item The attention region is not just the recognized object, but also some background as shown in Fig.10(b). This explains the precision values to be lower as compared to recall values.

\end{itemize}
     \begin{figure}[thpb]
      \centering
      \includegraphics{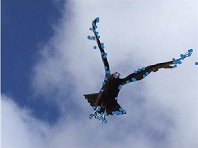}
      \caption{The important pixels tracked back after a forward pass through a VGG19 network.}
       \includegraphics{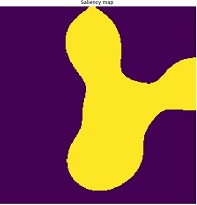}
      \caption{Saliency map derived for the picture shown above.}
      
       \includegraphics{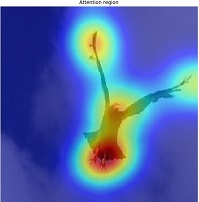}
      \caption{Attention region of the image based on the important pixel density. Red being the region with the most attention.}
      
      \label{figurelabel}
   \end{figure}
 \begin{figure}[thpb]
      \centering
    \includegraphics{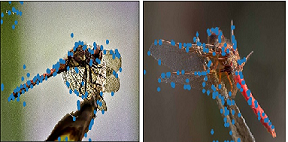}
      \caption{Results for the Dragonfly. Here there is a pattern that the CNN is directing most of it's attention at, the head and the tail of the dragonfly.}
        \label{figurelabel}
   \end{figure}

\begin{table}[h]
\caption{Results on MSRA-B dataset}
\label{table_example}
\begin{center}
\begin{tabular}{|c||c||c||c||c|}
\hline
Accuracy & Precision & Recall & F-Score & IOU\\
\hline
0.98 & 0.5 & 0.8 & 0.7 & 0.6\\
\hline
\end{tabular}
\end{center}
\end{table}
\begin{figure}[thpb]
      \centering
 \includegraphics{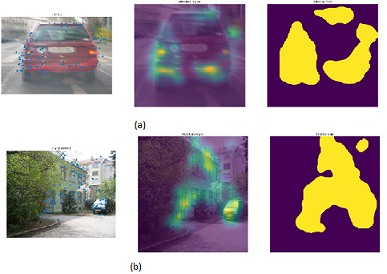}
      \caption{(a) Results on a blurred image of car. (b) Background is also shown some attention other than the class object car.}
        \label{figurelabel}
   \end{figure}

\section{FUTURE WORK}
In future we would like to extend this algorithm further on to densenet\cite{c12} and resnet\cite{c4}. We would also like to build an extension to architectures which stores the best activations directly and give a quick response. Also we would like to explore different applications as this can revolutionize the complexity.
\section{CONCLUSIONS}
We present a simple method to visualize and understand how a CNN looks at an image, by back tracking all the operations of CNN on an image. We have also shown that saliency parts in the image can be identified using this method. As shown in Fig.1. we have also visualized the attention region in the image. We would explore other applications in the future and also would like to improve this algorithm to make it faster and more generic. We succeeded in understanding a magnificent tool's working in a more simpler way and from a different point of view. 

\addtolength{\textheight}{-12cm}   




\section*{ACKNOWLEDGMENT}
I would like to express my special thanks of gratitude to my guide Dr.Deepak Mishra as well as Dr. Vineeth.B.S, who gave me the technical and moral support to do this wonderful project on the topic of Computer Vision, which also helped me in gaining a lot of knowledge and I came to know about so many new things I am really thankful to them.\par
Secondly I would also like to thank my parents and friends who helped me a lot in finalizing this project within the limited time frame.


\end{document}